\documentclass{article}
\usepackage{spconf,amsmath,graphicx}
\usepackage{bm}
\usepackage{bbm}
\usepackage{epsfig}
\usepackage{epstopdf}
\usepackage{subfigure}
\usepackage{color}
\definecolor{amethyst}{rgb}{0.6, 0.4, 0.8}
\definecolor{brandeisblue}{rgb}{0.0, 0.44, 1.0}
\definecolor{celadon}{rgb}{0.67, 0.88, 0.69}
\definecolor{crimson}{rgb}{0.86, 0.08, 0.24}
\definecolor{darkpastelgreen}{rgb}{0.01, 0.75, 0.24}
\definecolor{cyan(process)}{rgb}{0.0, 0.72, 0.92}
\usepackage{amsmath}
\usepackage[colorlinks,
            linkcolor=black,       
            anchorcolor=black,  
            citecolor=black,        
            ]{hyperref}

\usepackage[misc]{ifsym}

\title{Designing a 3D-aware StyleNeRF Encoder for Face Editing}
\name{Songlin Yang\textsuperscript{1,2}, Wei Wang\textsuperscript{2,*\thanks{* Corresponding author.} }, Bo Peng\textsuperscript{2}, Jing Dong\textsuperscript{2}}
\address{\textsuperscript{1}School of Artificial Intelligence, University of Chinese Academy of Sciences, Beijing, China\\
\textsuperscript{2}Center for Research on Intelligent Perception and Computing, NLPR, CASIA, Beijing, China\\
{\tt\small {yangsonglin2021@ia.ac.cn, $\{$wwang, bo.peng, jdong $\}$@nlpr.ia.ac.cn}}}
\begin{document}
\ninept
\maketitle
\begin{abstract}
GAN inversion has been exploited in many face manipulation tasks, but 2D GANs often fail to generate multi-view 3D consistent images. The encoders designed for 2D GANs are not able to provide sufficient 3D information for the inversion and editing. Therefore, 3D-aware GAN inversion is proposed to increase the 3D editing capability of GANs. However, the 3D-aware GAN inversion remains under-explored. To tackle this problem, we propose a 3D-aware (\textbf{3Da}) encoder for GAN inversion and face editing based on the powerful StyleNeRF model. Our proposed \textbf{3Da} encoder combines a parametric 3D face model with a learnable detail representation model to generate geometry, texture and view direction codes. For more flexible face manipulation, we then design a dual-branch StyleFlow module to transfer the StyleNeRF codes with disentangled geometry and texture flows. Extensive experiments demonstrate that we realize 3D consistent face manipulation in both facial attribute editing and texture transfer. Furthermore, for video editing, we make the sequence of frame codes share a common canonical manifold, which improves the temporal consistency of the edited attributes.
\end{abstract}
\section{Introduction}

\begin{figure}[htb]
        \centering
        \includegraphics[scale=0.3]{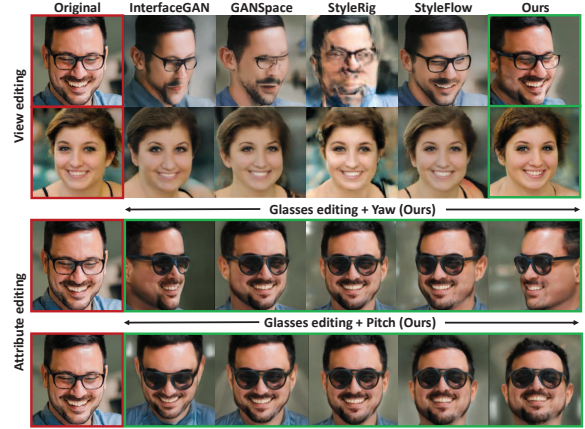}
        \caption{The comparisons of the multi-view editing effects, using StyleRig~\cite{tewari2020stylerig}, InterfaceGAN~\cite{shen2020interfacegan}, GANSpace~\cite{harkonen2020ganspace}, StyleFlow~\cite{abdal2021styleflow} and ours. Our method not only achieves better results in novel views, but also preserves the multi-view 3D consistency of the edited results.}
        \label{real_image_novel_view}
    \end{figure}

Face editing via GAN (Generative Adversarial Network) inversion~\cite{xia2022gan} enables users to flexibly edit a wide range of facial attributes in real face images. Existing methods~\cite{jiang2021talk,yin2022styleheat,abdal2022video2stylegan,tzaban2022stitch} first invert face images into the latent space of 2D GANs such as StyleGAN~\cite{karras2020analyzing}, then manipulate the style codes, and finally feed the edited codes into the pre-trained generator to obtain the edited face images. However, 2D GANs lack the knowledge of the underlying 3D structure of the faces, and their 3D consistency in multi-view generation is limited, as shown in Fig.~\ref{real_image_novel_view}.

In order to increase the 3D consistency of the generators in the GAN-inversion-based manipulation pipeline, one intuitive idea is replacing the 2D GANs with 3D-aware GANs~\cite{schwarz2020graf,niemeyer2021giraffe,gu2021stylenerf,chan2022efficient,xu20223d}. However, the vanilla encoders designed for 2D GAN inversion fail to provide sufficient 3D information for the 3D-aware GAN inversion. Furthermore, the SOTA 2D encoders like e4e~\cite{tov2021designing} bring much variety in the inversion stage, which degrades the video consistency in video editing. Therefore, to obtain better 3D consistency in multi-view facial attribute editing, we propose a 3D-aware (\textbf{3Da}) StyleNeRF~\cite{gu2021stylenerf} encoder which encodes geometry and texture separately to have more flexible manipulation capability. 

Our proposed \textbf{3Da} encoder combines a parametric 3D face model with a learnable detail representation model to generate the geometry, texture and view direction codes. By introducing the 3D face model, we can enhance the stability of the generated faces. The 3D-aware inversion codes are then fed into a well trained dual-branch StyleFlow~\cite{abdal2021styleflow} module which makes for the flexible face manipulation. We realize the 3D consistent face manipulation in both facial attribute editing and texture transfer. Moreover, we extend our pipeline to video editing. We make the video frames share a common canonical representation manifold, which improves the temporal consistency of the edited attributes.

The main contributions of this work are as follows: We propose the first 3D-aware (\textbf{3Da}) StyleNeRF encoder for the face editing. Our \textbf{3Da} encoder is able to encode geometry, texture and view direction information separately, achieving multi-view generation and facial attribute editing simultaneously. By introducing the parametric 3D face model, we are able to enhance the stability of the generated faces, which aligns the facial details with the morphable model adaptively.
    
\section{Method}
\begin{figure*}[htb]
        \centering
        \includegraphics[scale=0.53]{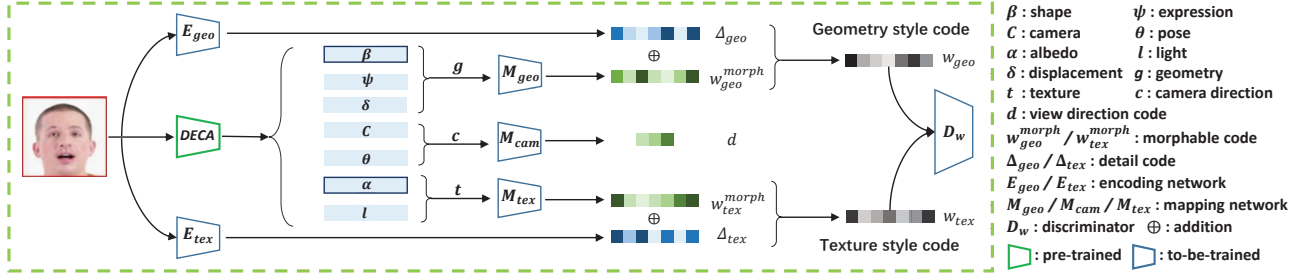}
        \caption{The framework of our 3D-aware (\textbf{3Da}) encoder. Note that when embedding the video frames, the shape $\bm{\beta}$ and albedo $\bm{\alpha}$ should be the same for the same face among different frames, i.e., frame-irrelevant. Therefore, in the video setting, we first extract these two coefficients for all the frames. Then we use the averaged $\bm{\beta}$ and $\bm{\alpha}$ for encoding all the frames.}
        \label{3Da}
    \end{figure*}
\begin{figure}[htb]
        \centering
        \includegraphics[scale=0.47]{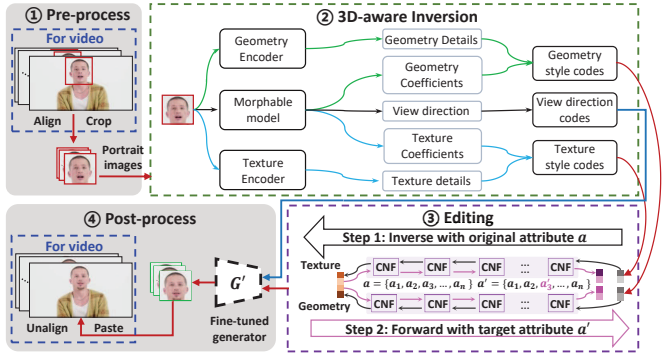}
        \caption{The portrait manipulation pipeline with our \textbf{3Da} StyleNeRF encoder for the image and video setting. Our 3D-aware GAN inversion module realizes the disentanglement of canonical and morphable modulations, as well as the separate editing of geometry and texture. Our attribute editing module extends the StyleFlow~\cite{abdal2021styleflow} to two branches for texture and geometry, respectively. Note that our \textbf{3Da} encoder is able to disentangle the frame-irrelevant information when encoding the video frame code sequence, which is beneficial to increase temporal consistency of the video editing.}
        \label{framework}
    \end{figure}

\subsection{3D-Aware StyleNeRF Inversion for Face Embedding}
The StyleNeRF~\cite{gu2021stylenerf} is adopted as our pre-trained generator. It has two inputs for conditioning the style and camera view respectively. Its NeRF-based~\cite{mildenhall2020nerf} architecture performs volume rendering only to produce a low-resolution feature map, and progressively applies upsampling in 2D to obtain high-resolution images. Our \textbf{first} motivation is that, two branches of geometry and texture should be adopted to match the 3D-aware architecture. The \textbf{second} motivation is the disentanglement of canonical and morphable information. So we adopt the parametric 3D face model DECA~\cite{feng2021learning} as 3D prior and use the ResNet-based~\cite{he2016deep} encoder to provide the detail information. This has two benefits: \textbf{(1)} For training, the sampled style codes with their corresponding synthetic images are used to train the encoder for encoding the images into the GAN space. Our proposed methods can accelerate the convergence and avoid overfitting to the synthetic data. \textbf{(2)} For the video setting, this can guide every code in the code sequence of video frames to share a canonical representation manifold of the target face, preserving temporal consistency of the inversion and editing.  

\begin{figure}[htb]
        \centering
        \includegraphics[scale=0.36]{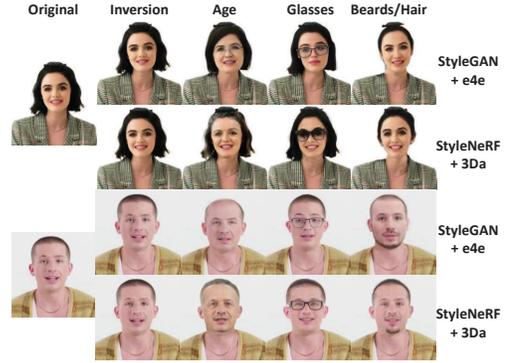}
        \caption{The edited face images of different attributes. Zoom in the digital version for better view.}
        \label{editing_results}
    \end{figure}
    
    \begin{figure}[htb]
        \centering
        \includegraphics[scale=0.26]{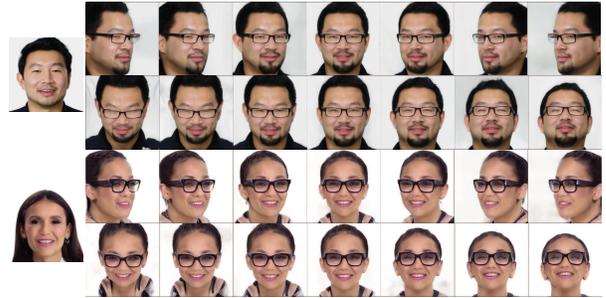}
        \caption{The multi-view generation of multi-attribute editing. }
        \label{novel_view}
    \end{figure}

\noindent
\textbf{Encoder.} As shown in Fig.~\ref{3Da}, for the geometry style code, $\bm{w_{geo}}=\bm{w_{geo}^{morph}}+\bm{\Delta_{geo}}$. For the texture style code, $\bm{w_{tex}}=\bm{w_{tex}^{morph}}+\bm{\Delta_{tex}}$. The addition operation combines the 3D information and content details. The CNN-based encoding networks $E_{geo}$ and $E_{tex}$ are used to extract the detail codes $\bm{\Delta_{geo}}$ and $\bm{\Delta_{tex}}$ respectively. For the morphable codes $\bm{w_{geo}^{morph}}$, $\bm{w_{tex}^{morph}}$ and view direction code $\bm{d}$, DECA~\cite{feng2021learning} is used to extract the semantic feature vectors of geometry $\bm{g}$, texture $\bm{t}$ and camera direction $\bm{c}$. Then, geometry $\bm{g}$ and texture $\bm{t}$ are input into fully-connected mapping networks $M_{geo}$ and $M_{tex}$ to obtain the morphable codes of geometry $\bm{w_{geo}^{morph}}$ and texture $\bm{w_{tex}^{morph}}$. Camera direction $\bm{c}$ is input into $M_{cam}$ to get the view direction codes $\bm{d}$. Specifically, geometry $\bm{g}$ is concatenated by shape $\bm{\beta}$ and expression $\bm{\psi}$ and displacement $\bm{\delta}$. Texture $\bm{t}$ is concatenated by albedo $\bm{\alpha}$, light $\bm{l}$. The camera direction $\bm{c}$ is concatenated by camera $\bm{C}$ and pose $\bm{\theta}$. 

\begin{figure}[htb]
        \centering
        \includegraphics[scale=0.30]{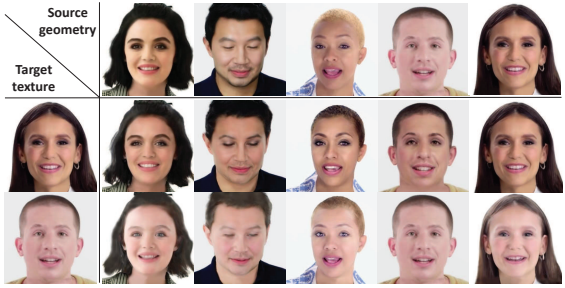}
        \caption{Texture transfer. The images in the first row provide the source geometry, and the images in the first column provide the target texture. Our \textbf{3Da} encoder is able to transfer the target texture to the source geometry.}
        \label{texture_transfer}
    \end{figure} 

\noindent
\textbf{Discriminator.} To encourage the style codes to lie within the distribution of the latent style code space of StyleNeRF, denoted as $\bm{W}$, a discriminator $D_{w}$ is used to discriminate between real samples from the $\bm{W}$ space and the learned latent space of our \textbf{3Da} encoder. This discriminator is important because it is able to not only accelerate convergence, but also avoid the mode collapse (see Fig.~\ref{ablation}).

\begin{table}[htb]
        \footnotesize
        \centering
       \begin{tabular}{ccccc}
        \hline Method&Age$\;\uparrow$&Glasses$\;\uparrow$&Beards$\;\uparrow$&Hair$\;\uparrow$\\
        \hline
        StyleGAN + e4e&0.637&0.653&0.651&0.694\\
        StyleNeRF + 3Da&\textbf{0.794}&\textbf{0.791}&\textbf{0.803}&\textbf{0.811}\\
        \hline
        \end{tabular}
        \caption{The identity consistency scores of edited face images.}
        \label{identity}
    \end{table}
    
\noindent
\textbf{Loss Function.} For formulation, we denote $E_{w}(\bm{x})=\{\bm{w_{i}}\}_{1\leq i\leq N}$ as style codes and $E_{d}(\bm{x})=\bm{d}$ as view direction codes, where $N$ is the number of style modulation layers ($N=21$ for StyleNeRF). The $E_{w}$ and $E_{d}$ represent the networks of our \textbf{3Da} encoder.  Note that each code in $\{\bm{w_i}\}_{1\leq i\leq 7}$ is the same and referred to as $\bm{w_{geo}}$, while each code in $\{\bm{w_i}\}_{8\leq i\leq 21}$ is the same and referred to as $\bm{w_{tex}}$. To optimize our encoder and discriminator in an adversarial manner, we use the non-saturating GAN loss function~\cite{goodfellow2014generative} to train these networks as follows:
    \begin{equation}
       \mathcal{L}_{adv}^{D,E}=-\mathop{\mathbbm{E}}\limits_{\bm{w}\sim W}[log D_{w}(\bm{w})]-\mathop{\mathbbm{E}}\limits_{\bm{x}\sim p_{X}}[log(1-D_{w}(E_{w}(\bm{x})))],
    \end{equation}
    \begin{equation}
        \mathcal{L}_{rec}^{E}=\mathcal{L}_{sim}+\lambda_{1} \mathcal{L}_{style}+\lambda_{2}\mathcal{L}_{view},
    \end{equation}
    where $L_{sim}$, $L_{style}$ and $L_{view}$ are as follows:
    \begin{equation}
         \mathcal{L}_{sim} =  \Vert \bm{x}-G(E_w(\bm{x}),E_d(\bm{x}))\Vert_{2}+vgg(\bm{x},G(E_w(\bm{x}),E_d(\bm{x}))),
    \end{equation}
    \begin{equation}
        \mathcal{L}_{style} = \Vert \bm{w_{geo}}-\bm{w_{geo}^{GT}} \Vert_{1}+\Vert \bm{w_{tex}}-\bm{w_{tex}^{GT}}\Vert_{1},
    \end{equation}
    \begin{equation}
        \mathcal{L}_{view}=\Vert \bm{d}-\bm{d^{GT}}\Vert_{1},
    \end{equation}

 The target image $\bm{x}$ is the style-mixing image with the ground-truth geometry style code $\bm{w_{geo}^{GT}}$, texture style code $\bm{w_{tex}^{GT}}$ and view direction code $\bm{d^{GT}}$. The $G$ is the fixed pre-trained generator. The $vgg$ denotes perceptual loss~\cite{johnson2016perceptual}. We set $\lambda_1$ and $\lambda_2$ as 0.5 and 5.

\begin{figure}[htb]
        \centering
        \includegraphics[scale=0.35]{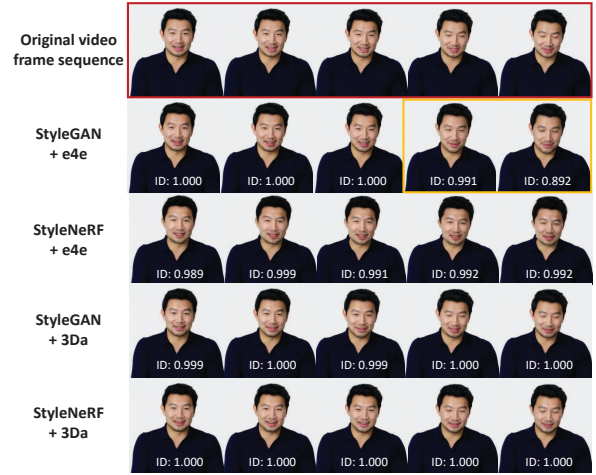}
        \caption{The inversion of video frame sequences. The StyleNeRF with \textbf{3Da} (fifth row) achieves better ID preservation results than others. The StyleGAN with e4e (second row) generates some inversion frames with low ID similarity (marked yellow). The e4e fails to keep the consistency of StyleNeRF, as shown in the third row.}
        \label{video_frame_inversion}
    \end{figure}
\subsection{Dual-Branch StyleFlow for Face Editing} We adopt StyleFlow~\cite{abdal2021styleflow} as the attribute editing method. However, the original StyleFlow only has a single branch of Continuous Normalizing Flow (CNF) blocks, failing to fully utilize the advantages of our \textbf{3Da} encoder. Therefore, as shown in the Fig.~\ref{framework}, we train two branches of Continuous Normalizing Flows $\{\phi_{s}\}_{s=geo,tex}$. Note that $\phi_{geo}$ and $\phi_{tex}$ are used to obtain geometry style code $\bm{w_{geo}}$ and texture style code $\bm{w_{tex}}$ respectively, for controllable editing. We denote $\bm{v}$ as the variable of the given StyleNeRF space, while $t$ is the time variable. We suppose that $\bm{w_{geo}}$ and $\bm{w_{tex}}$ are mapped from a latent variable $\bm z$ in a normal distribution. We use $\{\phi_{s}\}_{s=geo,tex}$ to conduct the inversion inference as follows:
    \begin{equation}
        \bm{v}(t_{0}) = \bm{v}(t_{1})+\int_{t_{1}}^{t_{0}} \phi_{s}(\bm{v}(t),t,\bm{a})dt,
    \end{equation}
where $\bm{v}(t_{0})$ is the $\bm{z}$. For $s=geo$, the $\bm{v}(t_{1})$ is $\bm{w_{geo}}$, while $\bm{v}(t_{1})$ is $\bm{w_{tex}}$ if $s=tex$. Note that $\bm{a}$ is the original attribute vector. Then, we modify $\bm{a}$ according to the given editing instruction, to obtain the edited attribute vector $\bm{a'}$. After that, we perform a forward inference to produce the edited style code $\bm{w'_{geo}}=\bm{v}(t_{1})$ or $\bm{w'_{tex}}=\bm{v}(t_{1})$, conditioned on $\bm{a'}$ as follows:
    \begin{equation}
        \bm{v}(t_{1}) = \bm{v}(t_{0})+\int_{t_{0}}^{t_{1}} \phi_{s}(\bm{v}(t),t,\bm{a'})dt,
    \end{equation}

The above is the inference process, and the training details of CNF blocks can be found in this work~\cite{abdal2021styleflow}.

\section{Experiments}
    
\subsection{Implementation Details}
\textbf{Network Architectures.} ResNet~\cite{he2016deep} is used as the backbone for encoding networks $E_{geo}$ and $E_{tex}$, to extract the feature vectors, corresponding to the input dimensions of StyleNeRF~\cite{gu2021stylenerf}. $M_{geo}$ and $M_{tex}$ are fully-connected networks with 5 layers, while $M_{cam}$ has 3 layers. The LeakyReLU is selected as the activation function. We conduct all the experiments on one NVIDIA RTX 3090. We conducted some preliminary prototyping using the MindSpore framework during our implementation. Our encoder requires 4 days, while the editing module requires 2 days.

\noindent
\textbf{Training Data and Annotation.} We randomly sample and save 10,000 groups of style-mixing style codes $\bm{w}$, view direction codes $\bm{d}$ and their corresponding StyleNeRF-generated images as training data. Moreover, the attribute vectors of these generated images are annotated using Microsoft Face API~\cite{microsoft2020api}, which every dimension of the vector represents an attribute. These attribute vectors and their corresponding style codes are used for training our dual-branch StyleFlow-based attribute editing module.

\begin{figure}[htb]
        \centering
        \includegraphics[scale=0.42]{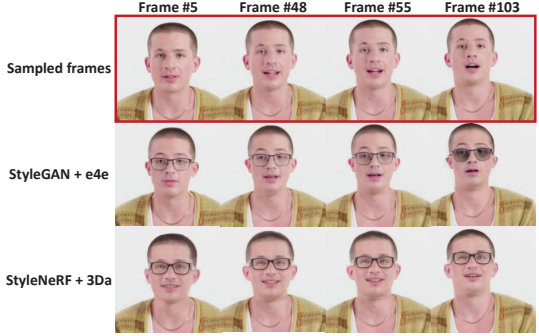}
        \caption{Comparison of the temporal consistency of video editing. The 'StyleGAN + e4e' method generates same changing degree of $Glasses$ leads to $Black~rimmed~glasses$ in the previous frames, while $Sun~glasses$ in the later frames. Our \textbf{3Da} with StyleNeRF is able to maintain the temporal consistency of the edited attributes.}
        \label{manipulation_consistency}
    \end{figure}
    
    \begin{table}[htb]
    \footnotesize
        \centering
        \begin{tabular}{ccccc}
        \hline
        Method  & PSNR$\;\uparrow$ & SSIM$\;\uparrow$ & VIF$\;\uparrow$ & FVD$\;\downarrow$\\
        \hline
        StyleGAN + e4e  & 40.3 & 0.996 & 0.75& 64.2\\
        StyleNeRF + e4e  & 32.1 & 0.995 & 0.71 & 168.1 \\
        StyleGAN + 3Da  & 38.4 & 0.995  & 0.78& 121.7\\
        StyleNeRF + 3Da  & \textbf{41.2} & \textbf{0.997} & \textbf{0.79}& \textbf{51.3}\\
        \hline
        \end{tabular}
        \caption{The quantitative evaluation of generation quality.}
        \label{quality}
    \end{table}

\noindent
\textbf{Baseline and Compared Methods.} The baseline of our experiments is the StyleGAN~\cite{karras2020analyzing} with e4e~\cite{tov2021designing} encoder that is wildly used in this field. Our differences are as follows: \textbf{(1)} Inputs: StyleNeRF has input of style codes and view direction codes, while StyleGAN has only style codes. \textbf{(2)} Style codes: The \textbf{3Da} encoder has only two different style codes, i.e., geometry and texture codes, while e4e output $N$ different style codes ($N=21$ or $N=18$ for StyleNeRF or StyleGAN). \textbf{(3)} Basic architecture: StyleNeRF adopts NeRF~\cite{mildenhall2020nerf} as basic generation networks, while StyleGAN lacks 3D prior. All the experiments are done under the same setting. We also compare with some other SOTA methods in the experiments, i.e., StyleRig~\cite{tewari2020stylerig}, InterfaceGAN~\cite{shen2020interfacegan}, GANSpace~\cite{harkonen2020ganspace}, StyleFlow~\cite{abdal2021styleflow}.

\noindent
\textbf{Metrics.} \textbf{\textit{Quality:}} PSNR, SSIM and VIF are used to measure the generation quality. Frechet Video Distance (FVD)~\cite{unterthiner2018towards} extends the FID~\cite{zhang2018unreasonable} to video quality settings. \textbf{\textit{Identity Consistency:}} We evaluate the identity of the edited faces using ArcFace~\cite{deng2019arcface} cosine similarity score. \textbf{\textit{Attribute Consistency:}} Following StyleFlow~\cite{abdal2021styleflow}, we use ResNet-18~\cite{he2016deep} trained on CelebA~\cite{liu2018large} as the facial attribute prediction model to output the attribute vectors of face images.

\subsection{Face Image Inversion and Editing}
\textbf{Attribute Editing.} As shown in Fig.~\ref{editing_results}, we select $Age$, $Glasses$, $Beards$ and $Hair$ as the examples. As shown in Tab.~\ref{identity}, we evaluate the identity consistency scores of edited face images compared with their original images in the FFHQ~\cite{karras2019style} dataset. The generation quality is quantitatively evaluated in Tab.~\ref{quality}.

\noindent
\textbf{Multi-Attribute Editing and Multi-View Generation.} As shown in Fig.~\ref{real_image_novel_view}, our method has good 3D consistency among different views of edited images. Furthermore, as shown in Fig.~\ref{novel_view}, our method can handle the multi-view generation and simultaneously edit multiple attributes.

\begin{figure}[htb]
        \centering
        \includegraphics[scale=0.38]{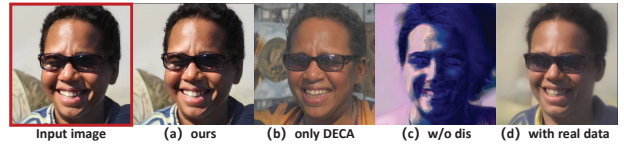}
        \caption{Ablation study of inversion results. \textbf{(a)} shows \textbf{3Da} encoder with discriminator and without real training data. \textbf{(b)} is mapping DECA coefficients to style codes. \textbf{(c)} is \textbf{3Da} encoder without the discriminator. \textbf{(d)} is \textbf{3Da} encoder with real data and the discriminator. Note that they are trained with the same epochs.}
        \label{ablation}
    \end{figure}

\noindent
\textbf{Texture Transfer.} As shown in Fig.~\ref{texture_transfer}, we can realize texture transfer among different real images by combing the geometry style code of one image with the texture style code of another and then inputting the style-mixing codes to the StyleNeRF. This illustrates the good geometry-texture disentanglement ability of our method.

\begin{table}[htb]
    \footnotesize
        \centering \begin{tabular}{ccccc}
        \hline Method&Age$\;\downarrow$&Glasses$\;\downarrow$&Beards$\;\downarrow$&Hair$\;\downarrow$\\
        \hline
        StyleGAN + e4e&0.498&0.533&0.502&0.497\\
        StyleNeRF + 3Da&\textbf{0.454}&\textbf{0.223}&\textbf{0.385}&\textbf{0.441}\\
        \hline
        \end{tabular}
        \caption{Temporal attribute inconsistency scores of video editing.}
        \label{attribute}
    \end{table}

\noindent
\subsection{Portrait Video Manipulation}
As shown in Fig.~\ref{framework}, our video manipulation pipeline is composed of three main stages, inspired by the STIT~\cite{tzaban2022stitch} method. First, we use DECA~\cite{feng2021learning} to extract the frame-irrelevant information (shape $\bm{\beta}$ and albedo $\bm{\alpha}$), encode the cropped face images and smooth the style code sequence over a window of two frames by weighted sum rules. Then, the cropped images and style code sequence are used to fine-tune the StyleNeRF generator. Note that the style code sequence is fixed in this fine-tuning process. And lastly, the style code sequence is input to our dual-branch StyleFlow to obtain an edited style code sequence conditioned on the required attribute vector, and the fine-tuned generator is used to obtain the edited frame sequence.

\noindent
\textbf{Video Inversion.} As shown in Fig.~\ref{video_frame_inversion}, we evaluate different methods under the same setting as STIT~\cite{tzaban2022stitch}. The StyleNeRF with our \textbf{3Da} encoder can achieve better results, in aspects of reconstruction and temporal consistency of identity similarity.

\noindent
\textbf{Video Editing.} As shown in Fig.~\ref{manipulation_consistency}, our method has more consistent video editing effects. As shown in Tab.~\ref{attribute}, we quantitatively measure the Mean Absolute Error of the attribute vectors between the first frame and the following frames on 50 testing videos. Each of them has 100 frames with different attribute edited. Our \textbf{3Da} encoder embeds the frame sequence more stably, as shown by the lower temporal attribute inconsistency.

\subsection{Ablation Study}
Fig.~\ref{ablation} (b) shows that only using DECA coefficients is insufficient. As shown in Fig.~\ref{ablation} (c) and Fig.~\ref{ablation} (d), \textbf{3Da} without the discriminator leads to the mode collapse, and using real data in training degrades the image quality. Using the real images without ground-truth style codes for training only has the $\mathcal{L}_{sim}$ loss, which is less effective than the embedding supervision from $\mathcal{L}_{style}$ and $\mathcal{L}_{view}$.
    
\section{Conclusion}
In this paper, we propose a 3D-aware (\textbf{3Da}) StyleNeRF encoder to encode geometry, texture and view direction of the real face images. Extensive experiments qualitatively and quantitatively demonstrate that, we are able to realize high-quality multi-view generation and facial attribute editing. Moreover, we extend our method to the portrait video manipulation, achieving better temporal consistency over the 2D-GAN-based editing methods.

\section{Acknowledgments}
This work is supported by the National Key Research and Development Program of China under Grant No. 2021YFC3320103, the National Natural Science Foundation of China (NSFC) under Grant 61972395, 62272460, a grant from Young Elite Scientists Sponsorship Program by CAST (YESS), and sponsored by CAAI-Huawei MindSpore Open Fund.

\bibliographystyle{IEEEbib}
\bibliography{refs}

\begin{thebibliography}{10}

\bibitem{xia2022gan}
Weihao Xia, Yulun Zhang, Yujiu Yang, Jing-Hao Xue, Bolei Zhou, and Ming-Hsuan
  Yang,
\newblock ``Gan inversion: A survey,''
\newblock {\em IEEE Transactions on Pattern Analysis and Machine Intelligence},
  2022.

\bibitem{jiang2021talk}
Yuming Jiang, Ziqi Huang, Xingang Pan, Chen~Change Loy, and Ziwei Liu,
\newblock ``Talk-to-edit: Fine-grained facial editing via dialog,''
\newblock in {\em Proceedings of the IEEE/CVF International Conference on
  Computer Vision}, 2021, pp. 13799--13808.

\bibitem{yin2022styleheat}
Fei Yin, Yong Zhang, Xiaodong Cun, Mingdeng Cao, Yanbo Fan, Xuan Wang, Qingyan
  Bai, Baoyuan Wu, Jue Wang, and Yujiu Yang,
\newblock ``Styleheat: One-shot high-resolution editable talking face
  generation via pretrained stylegan,''
\newblock {\em arXiv preprint arXiv:2203.04036}, 2022.

\bibitem{abdal2022video2stylegan}
Rameen Abdal, Peihao Zhu, Niloy~J Mitra, and Peter Wonka,
\newblock ``Video2stylegan: Disentangling local and global variations in a
  video,''
\newblock {\em arXiv preprint arXiv:2205.13996}, 2022.

\bibitem{tzaban2022stitch}
Rotem Tzaban, Ron Mokady, Rinon Gal, Amit~H Bermano, and Daniel Cohen-Or,
\newblock ``Stitch it in time: Gan-based facial editing of real videos,''
\newblock {\em arXiv preprint arXiv:2201.08361}, 2022.

\bibitem{karras2020analyzing}
Tero Karras, Samuli Laine, Miika Aittala, Janne Hellsten, Jaakko Lehtinen, and
  Timo Aila,
\newblock ``Analyzing and improving the image quality of stylegan,''
\newblock in {\em CVPR}, 2020, pp. 8110--8119.

\bibitem{tewari2020stylerig}
Ayush Tewari, Mohamed Elgharib, Gaurav Bharaj, Florian Bernard, Hans-Peter
  Seidel, Patrick P{\'e}rez, Michael Zollhofer, and Christian Theobalt,
\newblock ``Stylerig: Rigging stylegan for 3d control over portrait images,''
\newblock in {\em Proceedings of the IEEE/CVF Conference on Computer Vision and
  Pattern Recognition}, 2020, pp. 6142--6151.

\bibitem{shen2020interfacegan}
Yujun Shen, Ceyuan Yang, Xiaoou Tang, and Bolei Zhou,
\newblock ``Interfacegan: Interpreting the disentangled face representation
  learned by gans,''
\newblock {\em IEEE transactions on pattern analysis and machine intelligence},
  2020.

\bibitem{harkonen2020ganspace}
Erik H{\"a}rk{\"o}nen, Aaron Hertzmann, Jaakko Lehtinen, and Sylvain Paris,
\newblock ``Ganspace: Discovering interpretable gan controls,''
\newblock {\em Advances in Neural Information Processing Systems}, vol. 33, pp.
  9841--9850, 2020.

\bibitem{abdal2021styleflow}
Rameen Abdal, Peihao Zhu, Niloy~J Mitra, and Peter Wonka,
\newblock ``Styleflow: Attribute-conditioned exploration of stylegan-generated
  images using conditional continuous normalizing flows,''
\newblock {\em ACM Transactions on Graphics (ToG)}, vol. 40, no. 3, pp. 1--21,
  2021.

\bibitem{schwarz2020graf}
Katja Schwarz, Yiyi Liao, Michael Niemeyer, and Andreas Geiger,
\newblock ``Graf: Generative radiance fields for 3d-aware image synthesis,''
\newblock {\em Advances in Neural Information Processing Systems}, vol. 33, pp.
  20154--20166, 2020.

\bibitem{niemeyer2021giraffe}
Michael Niemeyer and Andreas Geiger,
\newblock ``Giraffe: Representing scenes as compositional generative neural
  feature fields,''
\newblock in {\em CVPR}, 2021, pp. 11453--11464.

\bibitem{gu2021stylenerf}
Jiatao Gu, Lingjie Liu, Peng Wang, and Christian Theobalt,
\newblock ``Stylenerf: A style-based 3d-aware generator for high-resolution
  image synthesis,''
\newblock {\em ICLR}, 2022.

\bibitem{chan2022efficient}
Eric~R Chan, Connor~Z Lin, Matthew~A Chan, Koki Nagano, Boxiao Pan, Shalini
  De~Mello, Orazio Gallo, Leonidas~J Guibas, Jonathan Tremblay, Sameh Khamis,
  et~al.,
\newblock ``Efficient geometry-aware 3d generative adversarial networks,''
\newblock in {\em CVPR}, 2022, pp. 16123--16133.

\bibitem{xu20223d}
Yinghao Xu, Sida Peng, Ceyuan Yang, Yujun Shen, and Bolei Zhou,
\newblock ``3d-aware image synthesis via learning structural and textural
  representations,''
\newblock in {\em Proceedings of the IEEE/CVF Conference on Computer Vision and
  Pattern Recognition}, 2022, pp. 18430--18439.

\bibitem{tov2021designing}
Omer Tov, Yuval Alaluf, Yotam Nitzan, Or~Patashnik, and Daniel Cohen-Or,
\newblock ``Designing an encoder for stylegan image manipulation,''
\newblock {\em ACM Transactions on Graphics (TOG)}, vol. 40, no. 4, pp. 1--14,
  2021.

\bibitem{mildenhall2020nerf}
Ben Mildenhall, Pratul~P Srinivasan, Matthew Tancik, Jonathan~T Barron, Ravi
  Ramamoorthi, and Ren Ng,
\newblock ``Nerf: Representing scenes as neural radiance fields for view
  synthesis,''
\newblock in {\em ECCV}. Springer, 2020, pp. 405--421.

\bibitem{feng2021learning}
Yao Feng, Haiwen Feng, Michael~J Black, and Timo Bolkart,
\newblock ``Learning an animatable detailed 3d face model from in-the-wild
  images,''
\newblock {\em ACM Transactions on Graphics (ToG)}, vol. 40, no. 4, pp. 1--13,
  2021.

\bibitem{he2016deep}
Kaiming He, Xiangyu Zhang, Shaoqing Ren, and Jian Sun,
\newblock ``Deep residual learning for image recognition,''
\newblock in {\em Proceedings of the IEEE conference on computer vision and
  pattern recognition}, 2016, pp. 770--778.

\bibitem{goodfellow2014generative}
Ian Goodfellow, Jean Pouget-Abadie, Mehdi Mirza, Bing Xu, David Warde-Farley,
  Sherjil Ozair, Aaron Courville, and Yoshua Bengio,
\newblock ``Generative adversarial nets,''
\newblock {\em Advances in neural information processing systems}, vol. 27,
  2014.

\bibitem{johnson2016perceptual}
Justin Johnson, Alexandre Alahi, and Li~Fei-Fei,
\newblock ``Perceptual losses for real-time style transfer and
  super-resolution,''
\newblock in {\em European conference on computer vision}. Springer, 2016, pp.
  694--711.

\bibitem{microsoft2020api}
Azure,
\newblock
  ``https://azure.microsoft.com/en-in/services/cognitive-services/face/,''
\newblock {\em Microsoft}, 2020.

\bibitem{unterthiner2018towards}
Thomas Unterthiner, Sjoerd van Steenkiste, Karol Kurach, Raphael Marinier,
  Marcin Michalski, and Sylvain Gelly,
\newblock ``Towards accurate generative models of video: A new metric \&
  challenges,''
\newblock {\em arXiv preprint arXiv:1812.01717}, 2018.

\bibitem{zhang2018unreasonable}
Richard Zhang, Phillip Isola, Alexei~A Efros, Eli Shechtman, and Oliver Wang,
\newblock ``The unreasonable effectiveness of deep features as a perceptual
  metric,''
\newblock in {\em Proceedings of the IEEE conference on computer vision and
  pattern recognition}, 2018, pp. 586--595.

\bibitem{deng2019arcface}
Jiankang Deng, Jia Guo, Niannan Xue, and Stefanos Zafeiriou,
\newblock ``Arcface: Additive angular margin loss for deep face recognition,''
\newblock in {\em CVPR}, 2019, pp. 4690--4699.

\bibitem{liu2018large}
Ziwei Liu, Ping Luo, Xiaogang Wang, and Xiaoou Tang,
\newblock ``Large-scale celebfaces attributes (celeba) dataset,''
\newblock {\em Retrieved August}, vol. 15, no. 2018, pp. 11, 2018.

\bibitem{karras2019style}
Tero Karras, Samuli Laine, and Timo Aila,
\newblock ``A style-based generator architecture for generative adversarial
  networks,''
\newblock in {\em CVPR}, 2019, pp. 4401--4410.

\end{thebibliography}

\end{document}